\documentclass[lettersize,journal]{IEEEtran}
\usepackage{amsmath,amsfonts}
\usepackage{algorithmic}
\usepackage{algorithm}
\usepackage{array}
\usepackage{color}
\usepackage[caption=false,font=footnotesize,labelfont=rm,textfont=rm]{subfig}
\usepackage{textcomp}
\usepackage{stfloats}
\usepackage{url}
\usepackage{verbatim}
\usepackage{graphicx}
\usepackage{cite}
\usepackage[colorlinks=true,linkcolor=black,citecolor=black,urlcolor=black]{hyperref}
\usepackage{epstopdf}
\usepackage{bm}
\usepackage{amsthm}
\usepackage{multicol}
\usepackage{multirow}
\usepackage{booktabs}

\DeclareMathOperator*{\Minimize}{min}
\begin{document}

\title{Learnable Residual-Based Latent Denoising in Semantic Communication\vspace{-0.07em}}

\author{
	Mingkai Xu, Yongpeng Wu, \IEEEmembership{Senior Member,~IEEE},\thanks{M. Xu, Y. Wu and W. Zhang are with the Department of Electronic Engineering, Shanghai Jiao Tong University, Shanghai 200240, China (e-mail: {michaelxu, yongpeng.wu, zhangwenjun}@sjtu.edu.cn).} Yuxuan Shi,\thanks{Y. Shi (\emph{corresponding author}) is with the School of Cyber and Engineering, Shanghai Jiao Tong University, Shanghai 200240, China (e-mail: ge49fuy@sjtu.edu.cn).} Xiang-Gen Xia, \IEEEmembership{Fellow,~IEEE},\thanks{X.-G. Xia is with the Department of Electrical and Computer Engineering, University of Delaware, Newark, DE 19716, USA (e-mail: xxia@ee.udel.edu).}\\ Wenjun Zhang, \IEEEmembership{Fellow,~IEEE}, Ping Zhang, \IEEEmembership{Fellow,~IEEE}\thanks{P. Zhang is with the State Key Laboratory of Networking and Switching Technology, Beijing University of Posts and Telecommunications, Beijing 100876, China (e-mail: pzhang@bupt.edu.cn).}}



\maketitle

\begin{abstract}
A latent denoising semantic communication (SemCom) framework is proposed for robust image transmission over noisy channels. By incorporating a learnable latent denoiser into the receiver, the received signals are preprocessed to effectively remove the channel noise and recover the semantic information, thereby enhancing the quality of the decoded images. Specifically, a latent denoising mapping is established by an iterative residual learning approach to improve the denoising efficiency while ensuring stable performance. Moreover, channel signal-to-noise ratio (SNR) is utilized to estimate and predict the latent similarity score (SS) for conditional denoising, where the number of denoising steps is adapted based on the predicted SS sequence, further reducing the communication latency. Finally, simulations demonstrate that the proposed framework can effectively and efficiently remove the channel noise at various levels and reconstruct visual-appealing images.
\end{abstract}

\begin{IEEEkeywords}
Semantic communication, image transmission, channel denoising, residual learning
\end{IEEEkeywords}

\section{Introduction}
\IEEEPARstart{S}{emantic} communication (SemCom) was envisioned by Shannon and Weaver to deliver the "semantic" meanings of data accurately~\cite{weaver1953recent}. Like other communication systems, SemCom also requires reliability to ensure the system mitigates the negative impacts of channel noise and interference during transmission. However, unlike classical information theory (CIT)-based systems where the primary goal is bit-error rate (BER) minimization, the goal of SemCom is to minimize semantic errors~\cite{weng2021semantic}. A semantic error is related to the transmission goal at the receiver, which usually demands the perceptual quality or the consistency of segmentation results~\cite{wang2022perceptual,grassucci2023generative} rather than pixel-wise minimal distortion.

To minimize a semantic error, it is crucial to counteract the impact of channel noise on the received semantic information. A few SemCom works have explored the denoising process in semantic transmission, which can be broadly divided into two categories: 1) \textbf{Latent denoising} that removes the noise from the channel received latents prior to the decoding process~\cite{cddm, witt,10437849}; 2) \textbf{Source denoising} that removes the noise from the reconstructed images after the decoding process~\cite{jiang2024diffsc,10829825}. However, the source denoising method enhances the decoded images in the high-dimensional pixel space, resulting in high computational complexity and low efficiency. In comparison, the latent denoising method is computationally less expensive, as it operates in a lower-dimensional latent space.

Moreover, many studies employ diffusion models (DMs)~\cite{NEURIPS2020_4c5bcfec} as their denoising networks, leveraging the powerful generative capacity of DM to reconstruct the distribution-preserved images. Nevertheless, the existing DM-based latent denoisers in SemCom still face two challenges. First, the amount of noise removed at each step is determined by a lengthy noise schedule, resulting in slow reverse sampling speed and large communication latency. Second, due to the generative characteristic, DM struggles to recover the semantic information accurately, which undermines the semantic fidelity of the latent. Specifically, in CDDM~\cite{cddm}, the number of sampling steps is related to the channel states, indicating that the latency could even increase as the channel states deteriorate; besides, it neglects the variance estimation in the sampling process, which leads to inaccurate latent recovery and suboptimal denoising performance.

To address these issues, instead of gradually removing the noise from the latent, we can boost the efficiency by directly learning the channel noise as a residual to remove it. In this letter, we propose a latent-level semantic denoising framework for image transmission, which consists of a transformer-based joint source-channel coding (JSCC) codec and a plug-in latent denoiser. Inspired by DnCNN~\cite{dncnn}, our denoiser adopts a single learnable residual mapping network to iteratively predict and remove the channel noise. This denoising process is conditioned on the similarity score (SS) which reflects the cosine similarity of the latents, thereby controlling the denoising intensity at each iteration. The SS is initialized based on the channel signal-to-noise ratio (SNR) and is dynamically updated by a learnable similarity predictor after each iteration. To better restore the image details, we promote the SS loss other than typical mean square error (MSE) loss during the network training. Furthermore, a denoising inference criterion is developed based on the sequence of predicted SS, enabling adaptive adjustment of the denoising steps to reduce the communication latency. Simulations verify the superiority of our method over separate source-channel coding (SSCC) and other JSCC methods, particularly in ultra-low SNR regimes.
\section{System Model and JSCC Architecture}
In this section, we describe the notational convenience, establish the system model of our framework, and introduce the architecture of the JSCC codec applied in our framework.

\subsection{Notational Convenience}
We denote scalars and vectors by lowercase letters in normal and bold fonts, respectively, e.g., scalar $x$ and vector $\bm{x}=(x_1,x_2,\cdots,x_L)^T$. $\mathbb{R}$ and $\mathbb{C}$ denote the sets of real and complex numbers, respectively. The identity matrix of size $k$ is denoted as $\boldsymbol{I}_k$. We define the distribution of an $n$-dimensional complex Gaussian random variable as $\mathcal{CN}(\bm{\mu},\bm{\Sigma})$ with mean vector $\bm{\mu}$ and covariance matrix $\bm{\Sigma}$. The mapping function is denoted by $f(\cdot; \bm{\varphi})$, with $\bm{\varphi}$ encapsulating all of its parameters. 

\begin{figure*}[tbp]
	\centering
	\includegraphics[width=0.79\textwidth]{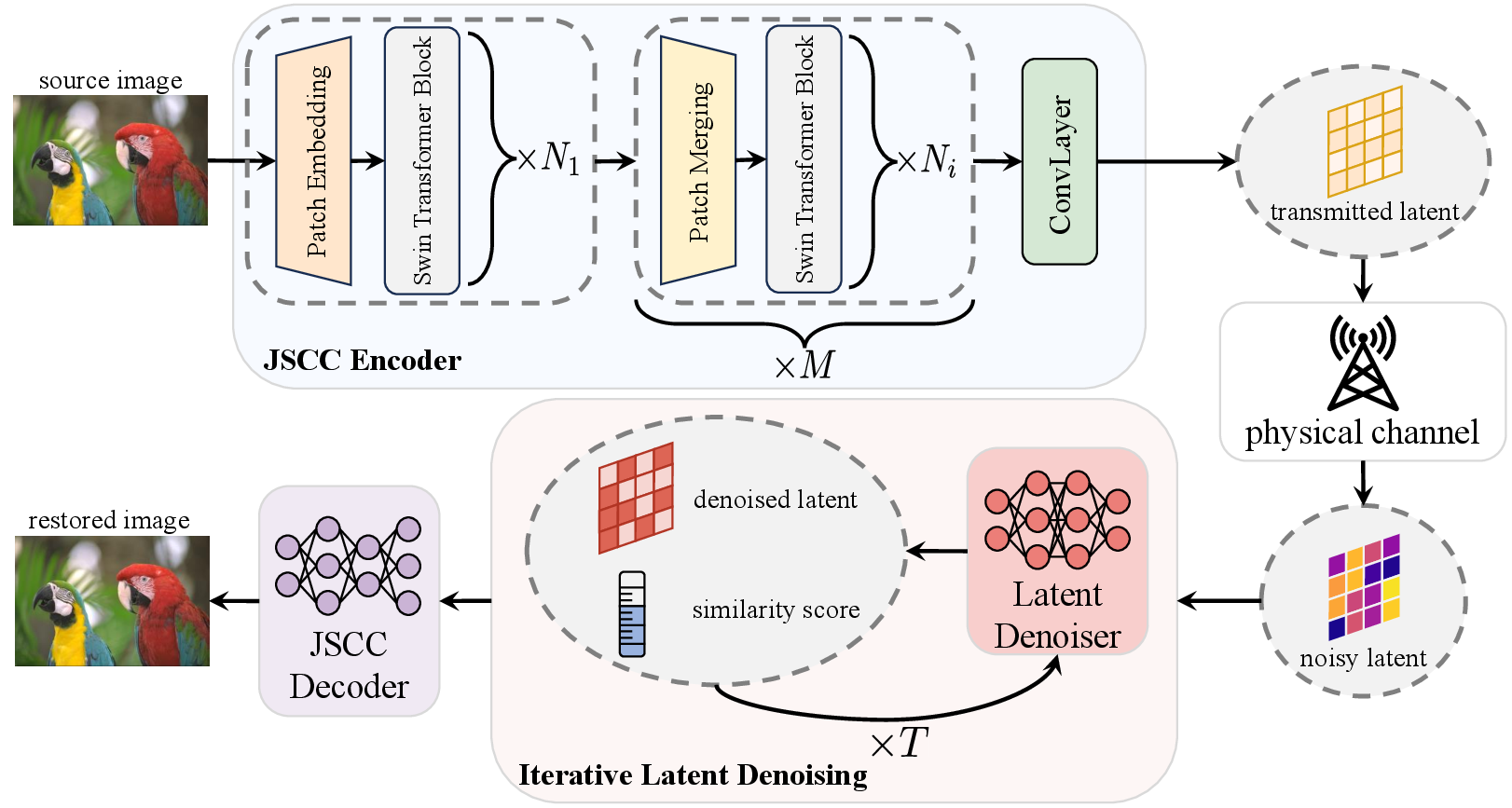}
	\caption{The overall framework of our proposed latent denoising and transmission scheme. In the diagram, the denoising process is iterated $T$ steps.}
	\label{fig:framework}
	
\end{figure*}

\subsection{System Model}
Consider that a source image $\bm{x}\in\mathbb{R}^k$ is transmitted over a noisy physical channel to the receiver for recovery. Specifically, we focus on the RGB image with the height $H$, the width $W$, and the channel number $3$, namely, $k=H\times W \times 3$.

Fig.~\ref{fig:framework} depicts our proposed image transmission and denoising framework. First, a JSCC encoder $\bm{y}=f_e(\bm{x}; \bm{\varphi})$ extracts the semantic information from the source image and produces the transmitted latent $\bm{y}\in \mathbb{R}^{2n}$. By combining two real-value symbols into one complex pair, $\bm{y}$ is converted to the channel input $\bm{y}_c\in \mathbb{C}^{n}$. Now, the \emph{channel bandwidth ratio} can be defined as $\rho=\frac{n}{k}$. After the power normalization operation~\cite{djscc}, $\bm{y}_c$ is transmitted over a physical channel. In this paper, we model a channel as the most common one with additive white Gaussian noise (AWGN):
\begin{equation}
	\bm{z}_c=\bm{y}_c+\bm{n}_c,
	\notag
\end{equation}
where $\bm{n}_c\sim\mathcal{CN}\left(\mathbf{0}, \sigma_n^2\boldsymbol{I}_n\right)$ denotes the AWGN with average power $\sigma_n^2$. The channel output $\bm{z}_c\in \mathbb{C}^{n}$ is then reshaped into the received noisy latent $\bm{z}_1\in \mathbb{R}^{2n}$. The proposed SS-conditioned latent denoiser is inserted between the channel and the JSCC decoder to remove the noise contained in $\bm{z}_1$. To ensure thorough noise removal, we perform iterative denoising with no more than $T_{\text{max}}$ steps. This iterative latent denoising process is formulated as $\hat{\bm{y}}=g_d(\bm{z}_1,\text{SNR}; \bm{\psi})$. Thereafter, the denoised latent $\hat{\bm{y}}$ is transformed by a JSCC decoder into the restored image, denoted by $\hat{\bm{x}}=f_d(\hat{\bm{y}};\bm{\theta})$. 

\subsection{JSCC Codec}
We choose Swin Transformer\cite{Liu_2021_ICCV} as the JSCC module of our framework due to its attention mechanism that identifies the most relevant aspects of the input data. The architecture of the JSCC encoder is detailed in Fig.~\ref{fig:framework}, which is composed of $M+1$ stages. The first stage contains one patch embedding block that divides an image into multiple patches, and $N_1$ swin transformer blocks that calculate the multi-head self-attention (MSA). Each of the subsequent $M$ stages contains one patch merging block and $N_i$ swin transformer blocks for the $i$-th stage. A convolution head is added at the end of the JSCC encoder to adjust the system's bandwidth consumption by varying the number of convolution filters. 
The JSCC decoder is also built by swin transformers and its structure resembles that of the JSCC encoder. 
The difference is that in the JSCC decoder, patch merging blocks are replaced with patch reverse merging blocks which perform up-sampling.

\section{Residual-based Iterative Latent Denoiser}
This section explains the working principle of our proposed latent denoiser. The architecture is introduced first, followed by the objective function, the proposed adaptive inference mechanism, and the training scheme.
\begin{figure}[tbp]
	\centering
	\includegraphics[width=0.49\textwidth]{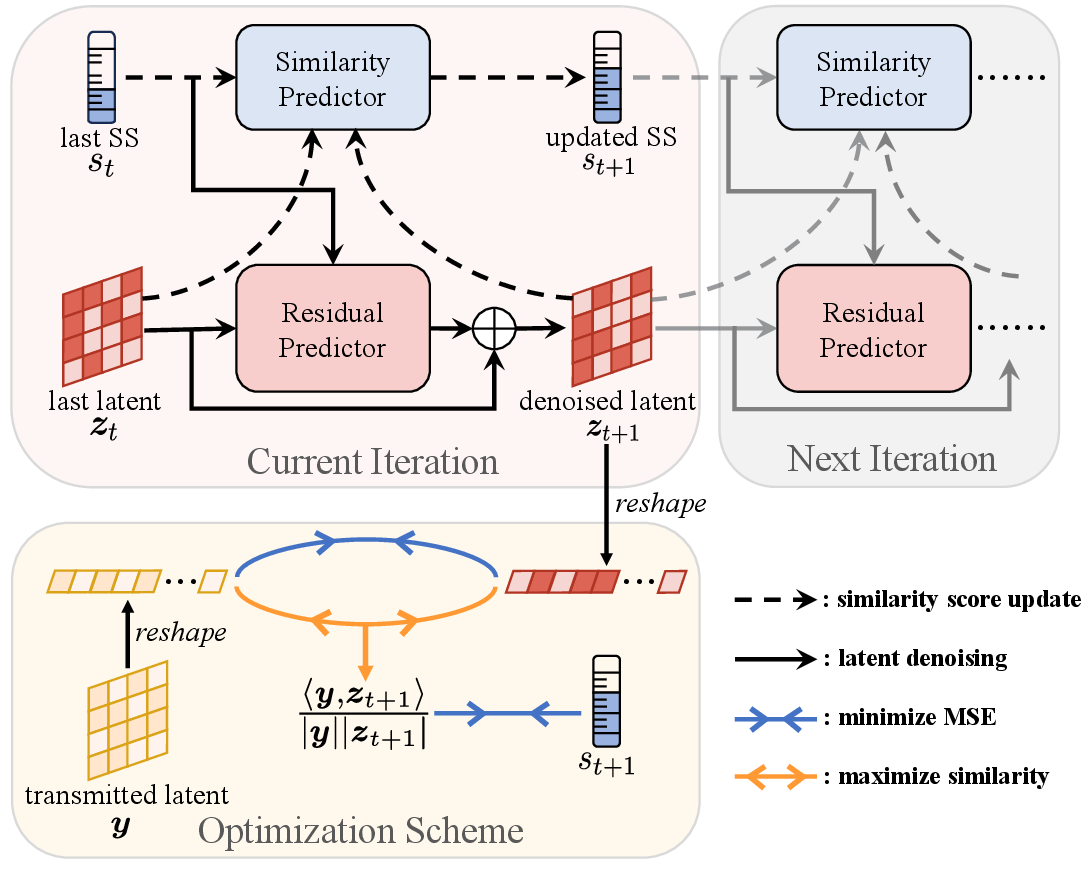}
	\caption{An illustration of our proposed denoising process and its optimization objective.}
	\label{fig:denoiser}
\end{figure}

\subsection{Working Principle}
Despite the contamination of channel noises, $\bm{z}_1$ still resembles $\bm{y}$ semantically. Therefore, the denoising mapping to learn is similar to the identity mapping. Hence, it is easier to optimize the residual mapping than the original denoising mapping~\cite{He_2016_CVPR}. Besides, utilizing the knowledge of channel noises can achieve a robust performance under varying SNRs.

Motivated by the above observation, we integrate the residual mapping with SNR-driven SS to establish our iterative latent denoiser, where SS measures the cosine similarity between the transmitted latent and the denoised latent. As shown in Fig.~\ref{fig:denoiser}, our denoiser consists of two neural networks: a residual predictor $g_r(\cdot,\cdot;\bm{\psi}_r)$ and a similarity predictor $g_s(\cdot,\cdot,\cdot;\bm{\psi}_s)$, with their parameters forming the complete set of denoiser parameters $\bm{\psi}=(\bm{\psi}_r,\bm{\psi}_s)$. $g_r$ predicts the residual latent conditioned on SS and then yields the denoised latent, while $g_s$ updates the SS to align it with the denoising progress. This process can be expressed as
\begin{align*}
	&\bm{z}_{t+1}=g_r(\bm{z}_t,s_t)+\bm{z}_t,\\
	&s_{t+1}=g_s(s_t,\bm{z}_t,\bm{z}_{t+1}),\quad t=1,2,\cdots,T_{\text{max}},
\end{align*}
where $z_{t+1}$ and $s_{t+1}\in \mathbb{R}$ represent the denoised latent in $t$-th iteration and its predicted SS, respectively. $\bm{r}_t=g_r(\bm{z}_t,s_t)$ denotes the predicted residual latent in $t$-th iteration. 

\begin{figure}[tbp]
	\centering
	\includegraphics[width=0.49\textwidth]{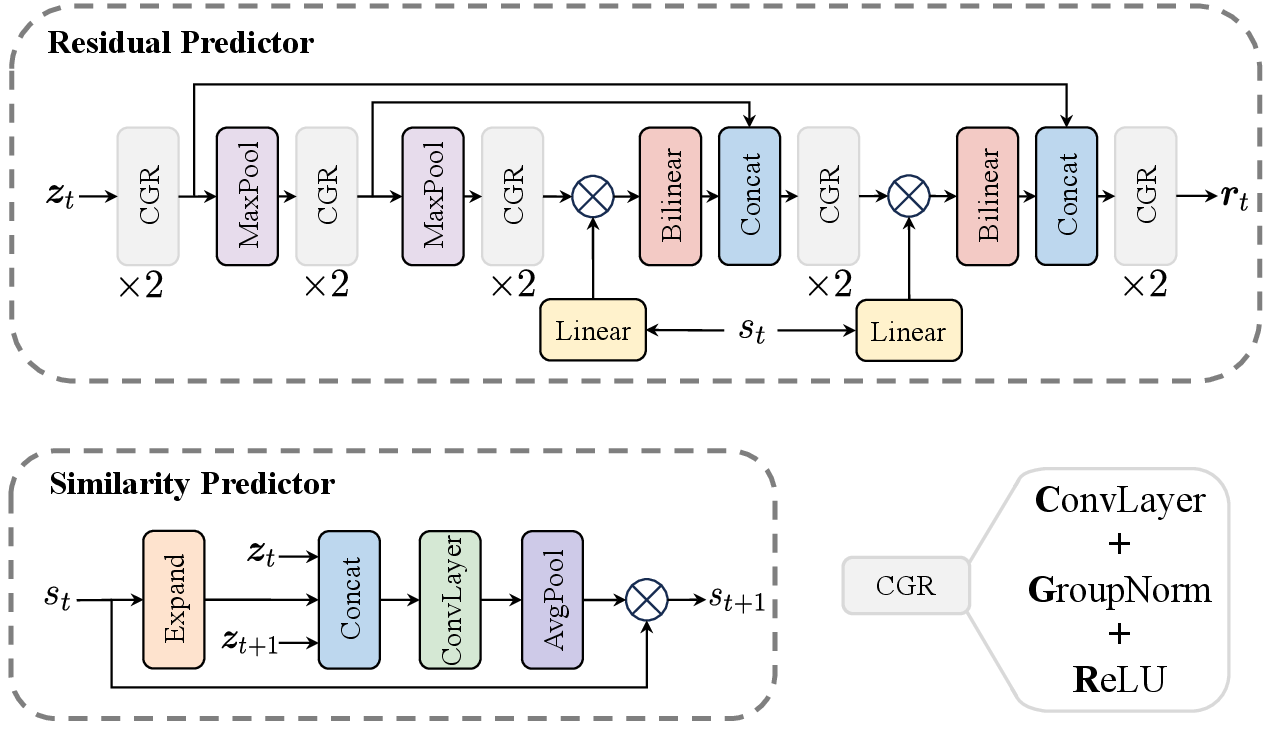}
	\caption{The architectures of the residual predictor and the similarity predictor.}
	\label{fig:denoise structure}
\end{figure}

Specifically, the structures of $g_r$ and $g_s$ are displayed in Fig.~\ref{fig:denoise structure}. For the residual predictor, we adopt a U-Net architecture due to the efficacy of downsampling U-Net stage in filtering high-frequency noises. A convolution layer, a group normalization layer, and a ReLU function are merged into a CGR block that processes the latent more effectively. During the residual prediction, $s_t$ is integrated with the upsampling process to guide the generation of the residual latent $\bm{r}_t$. As for the similarity predictor, we consider $(s_t,\bm{z}_t,\bm{z}_{t+1})$ as three joint inputs, which leverages the past similarity information to evaluate the current similarity accurately. 

Meanwhile, the value of $s_1$ is crucial as it impacts the performance of the first denoising step. Based on the definition of SS, $s_1$ should estimate the cosine similarity between the transmitted and received latents under AWGN channel. However, the closed-form solution is intractable due to its complicated distribution. Instead, we empirically initialize SS as
\begin{equation}
	 s_1=\frac{1}{\sqrt{1+\eta^{-1}}},\label{eq:s1}
\end{equation}
where $\eta$ denotes the channel SNR\footnote{We herein slightly abuse the notation of SNR. $\eta$ represents the ratio between the signal power and the noise power without conversion to dB.}. The effectiveness of such initialization will be validated in Sec.~\ref{sec:ablation}.

\subsection{Objective Function and Adaptive Inference Strategy}\label{sec:adaptive inference}
The training objective of the latent denoiser is to remove the channel noise in the received latent to retain the semantic information as much as possible under varying SNRs. We compute MSE between the transmitted and denoised latents to achieve this end, namely,
\begin{equation}
	\mathcal{L}_{\text{MSE}}=\mathbb{E}_{\bm{y},\bm{z}_t}\left[\frac{1}{2n}\Vert \bm{y}-\bm{z}_t \Vert_2^2\right],\quad t=2,3,\cdots,T_{\text{max}}.\notag
\end{equation}

However, MSE promotes pixel-wise average of the plausible latents which are typically overly-smooth~\cite{Ledig_2017_CVPR}. Furthermore, in low-SNR environments, merely applying MSE is insufficient to alleviate the channel noise pollution. Hence, we further apply the cosine similarity constraint to the denoised latent to facilitate residual learning under highly noisy conditions. The SS loss is computed as
\begin{align}
	&\mathcal{L}_{\text{SS}}=\mathbb{E}_{\bm{y},\bm{z}_t}\left[1-\frac{\bm{y}^T{\bm{z}_t}}{\Vert \bm{y} \Vert_2 \Vert \bm{z}_{t} \Vert_2}\right],\quad t=2,3,\cdots,T_{\text{max}}.\notag
\end{align}

Therefore, by combining the above two losses, the overall objective function of the residual predictor yields 
\begin{equation}
	\Minimize_{\bm{\psi}_r}  \mathcal{L}_{\text{MSE}}+\alpha\mathcal{L}_{\text{SS}},\label{eq:loss_gr}
\end{equation}
where $\alpha$ controls the trade-off between two loss items. The similarity predictor should estimate the SS accurately, and we apply MSE for the similarity prediction loss:
\begin{equation}
	\Minimize_{\bm{\psi}_s} \mathbb{E}_{s_t,\bm{y},\bm{z}_t}\left[\left(s_t-\frac{\bm{y}^T{\bm{z}_t}}{\Vert \bm{y} \Vert_2 \Vert \bm{z}_{t} \Vert_2}\right)^2\right],\quad t=2,3,\cdots,T_{\text{max}}.\label{eq:loss_gs}
\end{equation} 

Beyond this, we propose an adaptive inference approach to automatically adjust the denoising steps. In practice, excessive denoising steps may degrade the useful semantic information in the received latent, leading to a decline in overall performance. We validate the effectiveness of every step by analyzing the predicted SS. Specifically, suppose the inference of the latent denoiser produces the SS sequence $[s_1,s_2,\cdots,s_{T}]$. A monotonic non-decreasing sequence indicates that the denoising process constantly recovers more semantic information than it removes, and thus we regard these steps as effective. In this case, we repeat the iteration steps until it reaches $T_{\text{max}}$ or the SS decreases.

\subsection{Training Scheme for the Joint Latent Denoiser and JSCC}
The training of our framework is divided into three phases. First, we train the JSCC codec with the physical channel under the objective of end-to-end MSE, which is given by
\begin{equation}
\Minimize_{\bm{\varphi},\bm{\theta}} \mathbb{E}_{\bm{x},\hat{\bm{x}}}\left[\frac{1}{k}\lVert \bm{x}-\hat{\bm{x}} \rVert_2^2\right].\label{eq:e2e loss}
\end{equation}

Second, we freeze the weights of the pretrained JSCC encoder and insert the latent denoiser to be optimized by \eqref{eq:loss_gr} and \eqref{eq:loss_gs}. Finally, we concatenate all the modules and finetune only the weights of the JSCC decoder by \eqref{eq:e2e loss}, where $\bm{\varphi}$ is freezed and $\bm{\theta}$ is updated.

\section{Simulation Results}
In this section, numerical simulations are conducted to verify the image reconstruction and denoising performance of our proposed method under highly degraded channels. Besides, ablation studies and latency analysis are presented.
\subsection{Experimental Setup}
\subsubsection{Datasets}
In this paper, we choose the testing dataset to be Kodak24~\cite{kodak} which contains 24 RGB images with resolution $768\times512$, and the training dataset to be DIV2K~\cite{div2k} which includes 800 images with 2K resolution.

\subsubsection{Network and Training Details}
In the JSCC encoder, $M=3$ and $[N_1, N_2, N_3, N_4]=[2, 2, 6, 2]$. Except for the similarity predictor which uses a kernel size $1\times 1$ for the convolution layers, all other neural networks use a kernel size of $3\times 3$. For training, we set $\alpha=1$ in \eqref{eq:loss_gr} and apply Adam optimizer with the initial learning rate $10^{-4}$ decayed by poly strategy. The SNR is set 13dB in the first training phase and ranges from $\{0, 2, 4, 6, 8, 10\}$ in the second and third phases. 

\subsubsection{Baseline Schemes}
We consider the classical SSCC and learnable JSCC as the baseline schemes. For SSCC, BPG image codec~\cite{bpg} with low-density parity-check (LDPC) channel coding and 4-quadrature amplitude modulation (QAM) is included. The coding rates of LDPC are $0.75$ and $0.5$, denoted as \textbf{BPG+3/4+4Q} and \textbf{BPG+1/2+4Q}, respectively. For JSCC, we consider two transformer-based systems: \textbf{JSCC-T} built by removing the latent denoiser from our framework, and the \textbf{CDDM}. Each JSCC-based system is trained for $10^5$ iterations and we train one single system for the entire SNR regime.

\subsection{Image Reconstruction Performance under AWGN Channel}
\begin{figure}
	\centering
	\begin{minipage}[htbp]{0.63\linewidth}
		\centering
		\subfloat[]{
			\label{PSNR_SNR}
			\centering
			\includegraphics[width=1\textwidth]{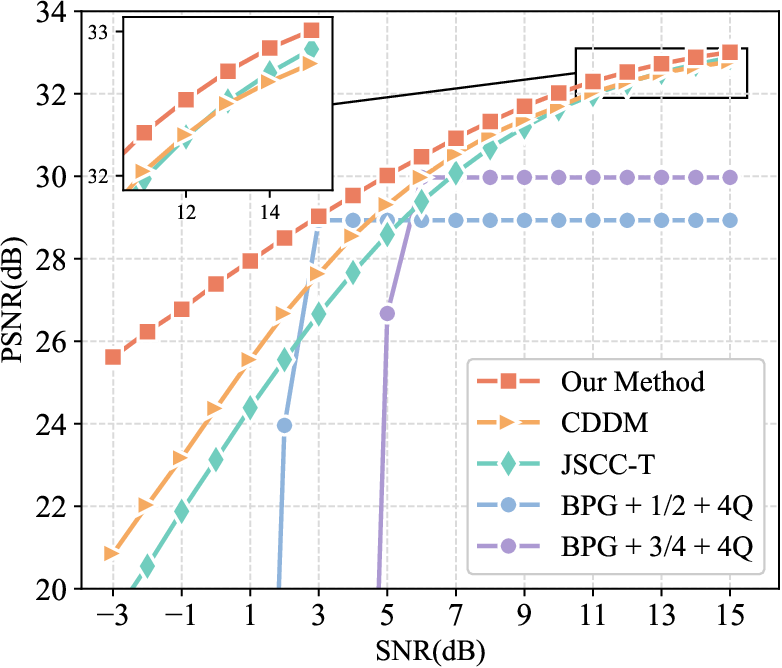}
		}
	\end{minipage}
	\begin{minipage}[tbp]{0.63\linewidth}
		\centering
		\subfloat[]{
			\label{SSIM_SNR}
			\centering
			\includegraphics[width=1\textwidth]{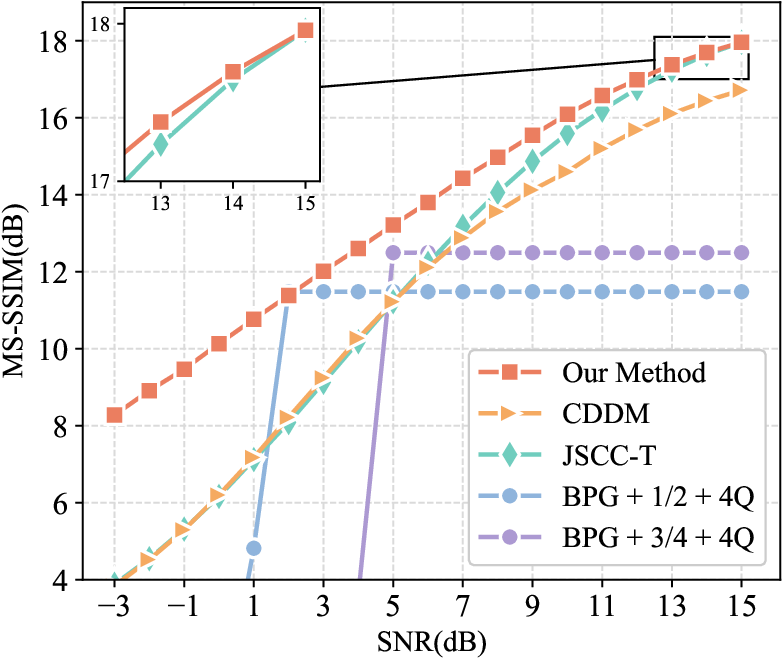}
		}
	\end{minipage}
	\caption{Comparison of reconstruction results of our method and the other baselines under AWGN channel. (a) PSNR versus SNR. (b) MS-SSIM versus SNR.}
	\label{fig:reconstruction performance}
	\vspace{-0.2cm}
\end{figure}
We evaluate the PSNR and MS-SSIM metrics of the images reconstructed by our method and different baselines under AWGN channel, where $\rho=\frac{1}{16}$. From Fig.~\ref{fig:reconstruction performance}\subref{PSNR_SNR} we observe that our method consistently provides PSNR gains, with the largest gain occurring under the worst channel condition (SNR=-3dB): $6.3$dB over JSCC-T and $4.7$dB over CDDM. This may be due to our framework not using a generative denoising method, which could otherwise lead to high-variance and unstable sampling in highly noisy channels. For MS-SSIM, Fig.~\ref{fig:reconstruction performance}\subref{SSIM_SNR} shows that under MSE optimization, JSCC-T and CDDM suffer from poor MS-SSIMs under low SNRs, and CDDM achieves little gain over JSCC-T, even becoming inferior when SNR$\textgreater 5$dB. In contrast, our method outperforms the other baselines under most SNRs. Therefore, our latent denoiser can be generalized to different metrics without specifically relying on them as the optimization objective. To sum up, this subsection verifies the ability of our latent denoiser to effectively remove extremely large channel noises.

\subsection{Ablation Study}\label{sec:ablation}
\begin{figure}[tbp]
	\centering
	\includegraphics[width=0.3\textwidth]{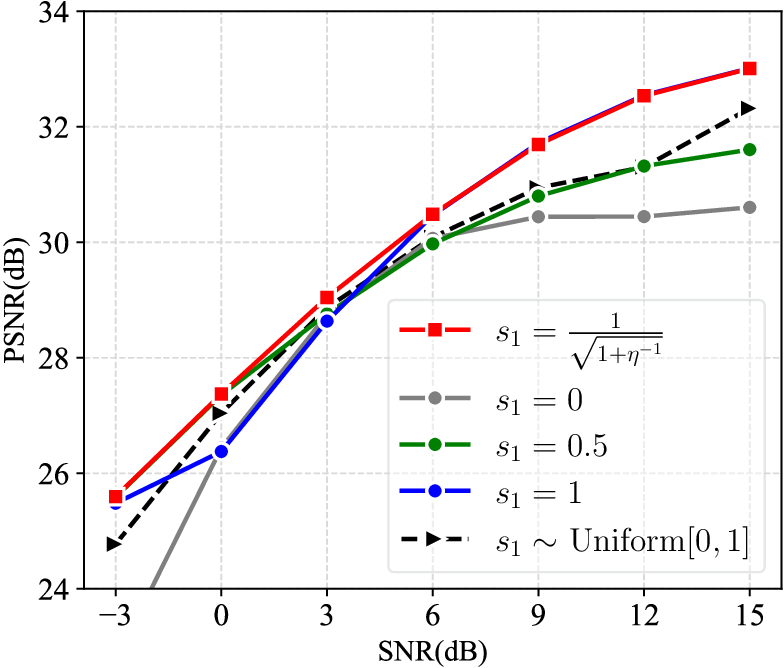}
	\caption{Comparisons of reconstruction results with different values of $s_1$.}
	\label{fig:ss_ablation}
\end{figure}

In this subsection, we first investigate different choices of the initial SS on the denoising performance. Specifically,  we compare our empirical initialization in~\eqref{eq:s1} with three deterministic initialization methods which select $s_1\in \{0, 0.5, 1\}$, and one random initialization method which samples $s_1$ uniformly from $[0, 1]$. As shown in Fig.~\ref{fig:ss_ablation}, our proposed choice consistently outperforms the other choices across the SNR regime.

\begin{figure}[!tbp]
	\centering
	\includegraphics[width=0.31\textwidth]{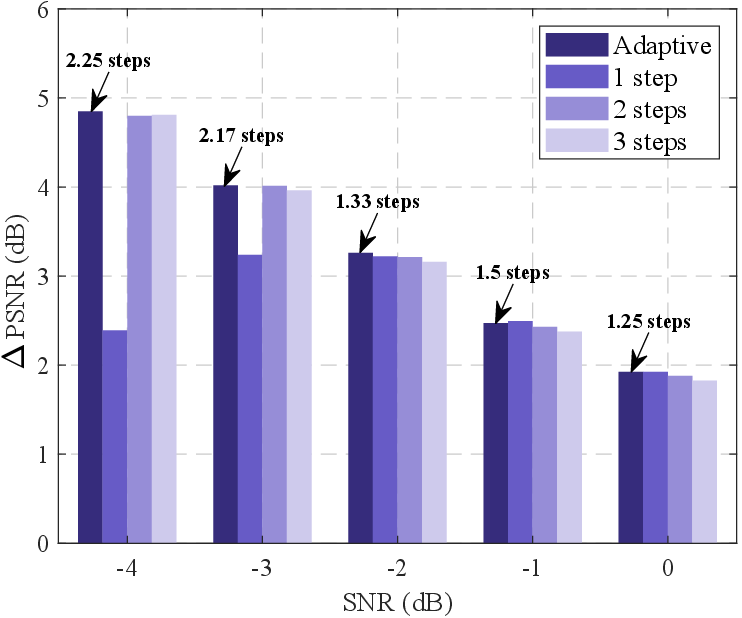}
	\caption{Comparison of the PSNR differences by varying the denoising steps.}
	\label{fig:ablation}
	\vspace{-0.1cm}
\end{figure}

Then, we evaluate our adaptive inference mechanism with $T_{\text{max}}$=3, varying the denoising steps and comparing the PSNR gains against no denoising. Fig.~\ref{fig:ablation} shows that one step at low SNR yields limited improvements while increasing steps at higher SNRs causes slight PSNR drops. Meanwhile, the adaptive inference provides near-optimal performance under SNR fluctuations while reducing the denoising steps, which further shortens the decoding time and becomes non-negligible when transmitting a large volume of images in practice.

\begin{figure}[htbp]
	{\small \text{\hspace{0.04\textwidth} \textbf{Original} \hspace{0.055\textwidth} \textbf{w/ SS} \emph{(29.63dB)} \hspace{0.022\textwidth} \textbf{w/o SS} \emph{(29.27dB)}}}\\
	\centering
	\begin{minipage}[t]{0.31\linewidth}
		\centering
		\vspace{-10pt}
		\subfloat{\includegraphics[width=1\textwidth]{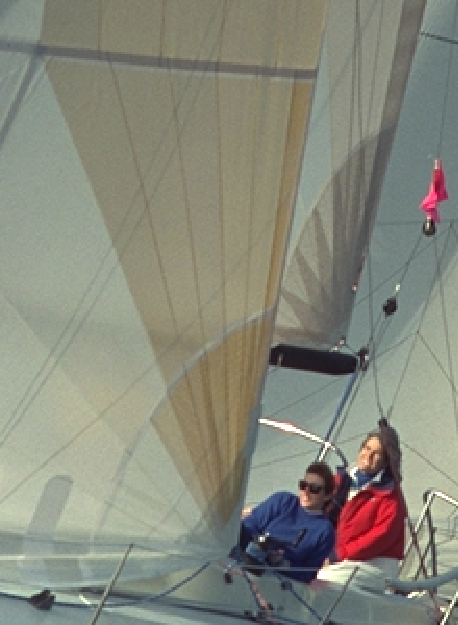}}
	\end{minipage}
	\begin{minipage}[t]{0.31\linewidth}
		\centering
		\vspace{-10pt}
		\subfloat{\includegraphics[width=1\textwidth]{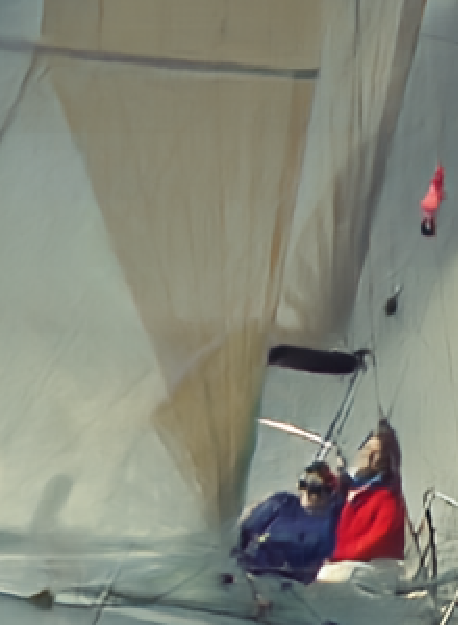}}
	\end{minipage}
	\begin{minipage}[t]{0.31\linewidth}
		\centering
		\vspace{-10pt}
		\subfloat{\includegraphics[width=1\textwidth]{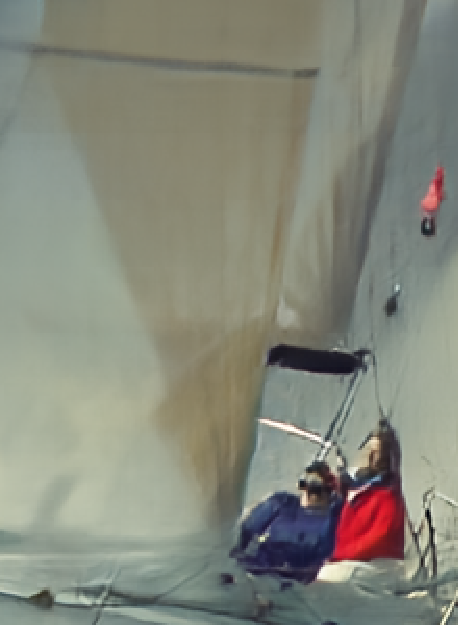}}
	\end{minipage}
	\par
	\vspace{7pt}
	\caption{Visualization of one image restored by our latent denoiser trained under different loss functions. SNR=0dB and PSNR values are annotated.}
	\label{fig:similarity ablation}
	\vspace{-0.1cm}
\end{figure}

To verify the impact of incorporating SS loss during training, we compare the reconstruction results with (w/) and without (w/o) SS loss. Fig.~\ref{fig:similarity ablation} shows that adding SS loss not only improves PSNR slightly but also recovers more high-frequency details (the wrinkles on the canvas) and avoids generating over-smooth images.

\subsection{Latency Analysis}
We compare the communication latency of different baselines, focusing on receiver processing latency. From Table.~\ref{Tab1} we can observe that compared with JSCC-T, CDDM incurs significant denoising latency even with increased SNR. In contrast, our denoising method does not add large latency but greatly enhances the transmission quality.

\begin{table}[htbp]
	\vspace{-0.2em}
	\renewcommand\arraystretch{0.9}
	\begin{center}
		\caption{Comparisons in terms of the latency of receivers in different image communication frameworks.\label{Tab1}}
		\begin{tabular}{ccccc}
			\toprule
			\multirow{2}{*}{Method} & \multicolumn{4}{c}{Latency (ms)} \\
			& SNR=-3dB & SNR=2dB & SNR=7dB & SNR=12dB \\
			\midrule
			JSCC-T & 23.12 & 22.78 & 25.05 & 24.19 \\	
			\textbf{Ours} & 45.16 & 37.02 & 38.85 & 41.55\\
			CDDM & 2106.94 & 2057.07 & 2052.14 & 1654.87 \\
			\bottomrule
		\end{tabular}
	\end{center}
	\vspace{-1.5em}
\end{table}
\section{Conclusion}
This work introduces a latent denoising SemCom framework that substantially improves the quality of the transmitted images over noisy channels. Through residual learning under MSE and SS optimizations for noise removal, the denoising efficiency is boosted and more image details are recovered. Furthermore, the SNR-based SS is leveraged to realize the conditional denoising and the adaptive inference, which guarantees robustness and low latency. Simulations show the superior performance of our method under fluctuating SNRs.

\bibliographystyle{IEEEtran}
\bibliography{IEEEabrv,ref}
\end{document}